\documentclass[a4paper,journal]{IEEEtran}

\usepackage{framed}
\usepackage{xcolor}
\usepackage{booktabs}
\usepackage{float}
\usepackage{graphicx}
\usepackage{multirow}
\usepackage{makecell}
\usepackage{float}
\usepackage{hyperref}
\usepackage{geometry}
\usepackage{titlesec}
\usepackage{amsmath,amsthm,amsfonts}
\usepackage{makecell}
\usepackage[numbers,sort&compress]{natbib}
\usepackage[ruled,vlined]{algorithm2e}

\usepackage{subcaption}
\usepackage{tikz}
\usepackage{siunitx}
\usepackage{tcolorbox}
\titlespacing*{\section}{0pt}{1.3ex}{1.3ex}
\titlespacing*{\subsection}{0pt}{0.9ex}{0.9ex}

\usepackage{tikz,xcolor}
\definecolor{lime}{HTML}{A6CE39}
\DeclareRobustCommand{\orcidicon}{
    \begin{tikzpicture}
        \draw[lime, fill=lime] (0,0)
        circle[radius=0.13]
        node[white]{{\fontfamily{qag}\selectfont \tiny \.{I}D}};
    \end{tikzpicture}
    \hspace{-2mm}
}
\foreach \x in {Zhuoyi, Haoyu, Zhen, Ershun}{\expandafter\xdef\csname orcid\x\endcsname{\noexpand\href{https://orcid.org/\csname orcidauthor\x\endcsname}{\noexpand\orcidicon}}
}

\begin{document}

\title{Large Language Model-Empowered Intelligent Agent for Adaptive Fast-Charging Control of Lithium-Ion Batteries}
\author{
    Joey Chan\hspace{-2mm}\orcidZhuoyi{}\hspace{-1mm},
    Haoyu Pan\hspace{-2mm}\orcidHaoyu{}\hspace{-1mm},~\IEEEmembership{Student Member,~IEEE},\\
    Zhen Chen\hspace{-2mm}\orcidZhen{}\hspace{-1mm},~\IEEEmembership{Member,~IEEE},
    Ershun Pan\hspace{-2mm}\orcidErshun{}\hspace{-1mm},~\IEEEmembership{Senior Member,~IEEE}

    \thanks{Joey Chan, Haoyu Pan, Zhen Chen (Coresponding Author), and Ershun Pan are with the Department of Industrial Engineering and Management, Shanghai Jiao tong University, Shanghai, China.
    }
    }

\maketitle
\vspace{-7cm} 

\begin{abstract}
Fast charging of lithium-ion batteries is fundamentally a multi-objective control problem that requires reducing charging time and sustaining high energy throughput while suppressing accelerated degradation and safety-critical electrochemical reactions. Conventional charging protocols are straightforward to implement but have limited capability to adapt to variations in battery chemistry, operating condition, and aging state, whereas model-based optimization and conventional data-driven control often rely on accurate models, expensive iterative search, or task-specific policies. In this work, we develop a large language model (LLM)-empowered intelligent agent for adaptive battery fast-charging control. Rather than using the LLM merely as a predictor, the proposed framework formulates battery charging as a closed-loop interaction process in which the agent interprets structured battery states, generates executable charging actions, observes the resulting battery response, and continuously adjusts subsequent decisions. A large-scale simulation corpus is constructed using PyBaMM across five battery chemistries and 11 parameter sets, and the resulting state--action--response trajectories are transformed into structured process data for agent training. We further propose a two-stage supervised fine-tuning and Group Relative Policy Optimization (SFT--GRPO) strategy. SFT first establishes battery-control semantics and feasible decision priors from simulation-derived demonstrations, after which GRPO aligns the agent with a multi-objective control objective involving energy throughput, degradation mitigation, and operational constraints. Closed-loop simulation experiments demonstrate that, under matched energy-throughput requirements, the proposed agent achieves longer cycling life than conventional charging baselines and optimization-based strategies across heterogeneous battery configurations. This study provides a simulation-grounded proof of concept for using LLM agents as sequential decision makers in battery management and represents a step toward language-model-enabled control of physical energy systems.
\end{abstract}


    \section{Introduction}

Lithium-ion batteries (LIBs) have become a key energy-storage technology for electric vehicles and grid-scale energy-storage systems because of their high energy density, high efficiency, and long cycle life\cite{chan2026universal}. 
With the rapid growth of electrified transportation and renewable-energy integration, fast charging has become increasingly important\cite{tomaszewska2019lithium}. 
However, high charging currents also intensify electrochemical polarization and heat generation and may accelerate solid-electrolyte-interphase (SEI) growth, lithium plating, and other degradation mechanisms\cite{park2020reinforcement,ji2025design}. 
These effects vary with the state of charge (SOC), temperature, state of health (SOH), battery chemistry, and operating history. 
Fast charging is therefore fundamentally a multi-objective control problem that requires balancing charging speed and energy throughput against degradation and safety\cite{jiang2025fast}.

Conventional charging strategies, particularly constant-current--constant-voltage (CC--CV) and multistage constant-current (MSCC) protocols, are widely adopted because of their simplicity and ease of implementation\cite{trimboli2015charging}. 
Nevertheless, their control parameters are generally determined offline and remain largely fixed during operation. 
As a battery ages, its capacity, resistance, polarization characteristics, and degradation boundaries continuously evolve, meaning that a protocol optimized for a fresh cell or a particular battery configuration may no longer remain suitable under different aging states or chemistries\cite{gao2019health}. 
This limitation has motivated the development of adaptive and health-aware charging strategies.

Model-based optimal charging represents one major research direction. Equivalent circuit models (ECMs) \cite{liu2018multi,liu2017constrained} provide computationally efficient descriptions of battery dynamics, whereas electrochemical models \cite{guo2011single,kong2020pseudo} enable explicit consideration of internal physical states and degradation mechanisms. 
For example, Gao et al.\ \cite{gao2019health} developed a health-aware strategy based on an electrochemical--thermal--aging model, while optimized MSCC strategies have been proposed to constrain polarization and thermal effects \cite{zhang2017charging}. 
Despite their physical interpretability, these approaches face a trade-off between model fidelity and computational efficiency. 
Their control performance also depends strongly on model and parameter accuracy, while repeated solution of constrained optimization problems can become computationally demanding under varying battery conditions.

Data-driven optimization provides an alternative by identifying high-performance charging protocols directly from experimental or simulation results. 
Bayesian optimization has been particularly effective in reducing the number of evaluations required to search large protocol spaces. 
Attia et al.\ \cite{attia2020closed}, for example, combined early cycle-life prediction with Bayesian optimization to accelerate the discovery of fast-charging protocols, and related studies have further demonstrated the potential of data-driven protocol optimization \cite{jiang2022fast}. However, such approaches primarily search for high-performing protocols within predefined spaces. As the number of control variables and battery conditions increases, the search space grows rapidly, and additional optimization may be required when the battery configuration or operating objective changes.

Reinforcement learning (RL) further shifts battery charging from offline protocol search toward sequential decision making\cite{zhao2024reinforcement}. 
An RL controller learns a mapping from battery states to charging actions and improves the policy according to feedback involving charging time, degradation, temperature, energy efficiency, or other objectives \cite{zhang2025fast,hao2023adaptive}. 
Nevertheless, most existing RL-based charging methods employ task-specific neural-network policies constructed for predefined observation spaces, action spaces, and battery environments. 
Their adaptation across heterogeneous chemistries, parameter configurations, and operating requirements therefore remains challenging.

Recent advances in large language models (LLMs) and LLM-based agents provide a new perspective on this problem. 
Domain-specific models such as BloombergGPT \cite{wu2023bloomberggpt}, Minerva \cite{lewkowycz2022solving}, and Llemma \cite{azerbayev2024llemma} have demonstrated the ability of pretrained language models to acquire specialized knowledge and reasoning capabilities.
Meanwhile, agentic frameworks such as ReAct \cite{yao2022react} and Voyager \cite{wang2023voyager} have extended LLMs from passive text generation toward interactive decision making, where models interpret environmental information, generate executable actions, and adjust their behavior according to feedback.

Battery charging naturally possesses several characteristics compatible with this paradigm. 
Battery systems continuously provide structured numerical information describing their operating and health states, while charging commands can be expressed as structured and machine-executable actions. 
More importantly, charging is a long-horizon problem in which short-term energy gains must be balanced against degradation accumulated over repeated cycles. 
An LLM-based agent may therefore provide a flexible decision architecture capable of integrating heterogeneous battery states, operating history, control constraints, and long-term objectives within a common representation.
A fundamental challenge is the acquisition of sufficient interaction data for training such an agent. Large-scale exploration directly on physical batteries is costly and time-consuming, because degradation experiments typically span hundreds or thousands of cycles and aggressive trial-and-error control may expose cells to undesirable conditions\cite{long2024llms}. 
Physics-based numerical simulation provides an alternative means of generating diverse and repeatable battery trajectories while retaining information about internal electrochemical and degradation states.

In this work, we use Python Battery Mathematical Modelling (PyBaMM) \cite{sulzer2021python} as the simulation environment and construct large-scale charging-control data covering five battery chemistries and 11 parameter sets. 
The resulting battery states, charging actions, and responses are organized into structured interaction trajectories for LLM training. Rather than using the LLM as a one-shot protocol generator, we formulate it as the decision core of a battery charging agent that adjusts charging actions according to the evolving battery state and feedback from the simulation environment.
To train the agent, we propose a two-stage supervised fine-tuning and Group Relative Policy Optimization (SFT--GRPO) framework. 
SFT first learns battery-control knowledge, feasible action structures, and decision patterns from high-quality simulation trajectories. 
The fine-tuned model is then further optimized using GRPO \cite{shao2024deepseekmath}, where candidate actions are evaluated according to a multi-objective reward involving energy throughput, battery degradation, and operational constraints. 
In this way, SFT provides an initial control prior, while GRPO enables the agent to improve its policy according to the physical outcomes of its decisions.

The resulting framework separates physical state evolution from decision making: PyBaMM provides the physics-based battery environment, while the LLM learns the control policy. 
This design allows extensive virtual interaction without requiring large-scale trial-and-error experiments on physical cells and provides a practical foundation for investigating agentic control of battery systems.
The main contributions of this work are summarized as follows:

\begin{enumerate}

\item \textbf{A simulation-grounded framework is developed for LLM-based battery charging control.}
A large-scale simulation corpus spanning five physics-based battery models and 11 parameter sets is constructed using PyBaMM. Heterogeneous battery trajectories are transformed into structured state--action data, enabling the LLM to acquire battery-control knowledge from diverse electrochemical dynamics while reducing reliance on costly physical experiments.

\item \textbf{A two-stage SFT--GRPO strategy is proposed for physics-grounded charging optimization.}
SFT establishes structured action-generation capabilities and feasible control priors through expert demonstrations, while GRPO improves charging decisions using physics-evaluated feedback on SOC progress, degradation, temperature, and operational constraints. Combined with a deterministic safety filter, the framework enables adaptive, closed-loop charging control through executable LLM-generated actions.

\item \textbf{Cross-configuration evaluations demonstrate improved capacity retention and charging efficiency.}
Extensive simulations across four battery model--parameter-set combinations under isothermal and non-isothermal conditions demonstrate the effectiveness of the proposed method. Compared with conventional and learning-based charging strategies, the LLM achieves superior capacity retention in most configurations under matched energy throughput while maintaining competitive charging efficiency, highlighting its potential for degradation-aware battery management.

\end{enumerate}

The present study focuses on simulation-based development and validation. 
The structured observation and action interfaces nevertheless provide a pathway toward future hardware-in-the-loop and real-cell implementation, where the simulation environment can be replaced by a battery cycler or battery-management system.
The remainder of this paper is organized as follows. 
Section~2 formulates the fast-charging problem and introduces the simulation environment and training data. 
Section~3 presents the proposed LLM-based charging agent and SFT--GRPO framework. 
Section~4 evaluates its control performance. Section~5 discusses model behavior and deployment considerations. 
Finally, Section~6 concludes the paper.

    \section{Battery Simulation Corpus Generation}
\label{sec:battery-simulation-corpus}

To expose the language model to battery dynamics that are physically
consistent but sufficiently diverse for policy learning, we generated synthetic
charging trajectories with the Python Battery Mathematical Modelling (PyBaMM) framework~\cite{sulzer2021python}.  
The corpus spans two battery chemistries and five physics-based model families.  

\subsection{Physics-Based Battery Models}
\label{subsec:physics-based-models}

Table~\ref{tab:battery-models} summarizes the five model families.  The
equations shown are the defining or representative relations rather than the
complete boundary-value problems.  For the lithium-ion models, \(c_{s,k}\) is
the solid-phase lithium concentration in electrode \(k\in\{n,p\}\), \(c_e\)
is the electrolyte concentration, \(\phi_s\) and \(\phi_e\) are the
solid- and electrolyte-phase potentials, \(j_k\) is the interfacial reaction
current density, and \(U_k\) is the open-circuit potential.  The radial
coordinate inside an idealized active-material particle is \(r\), whereas
\(x\) denotes the through-cell coordinate.  The symbols \(D_{s,k}\),
\(D_e^{\mathrm{eff}}\), \(\sigma_k^{\mathrm{eff}}\), and
\(\kappa^{\mathrm{eff}}\) denote effective transport coefficients.  The
same conservation-law notation is used for the lead--acid models, with
\(\varepsilon\) denoting porosity and \(N_e\) the electrolyte molar flux.

\begin{figure*}[htbp] 
  \centering
  \includegraphics[width=\linewidth]{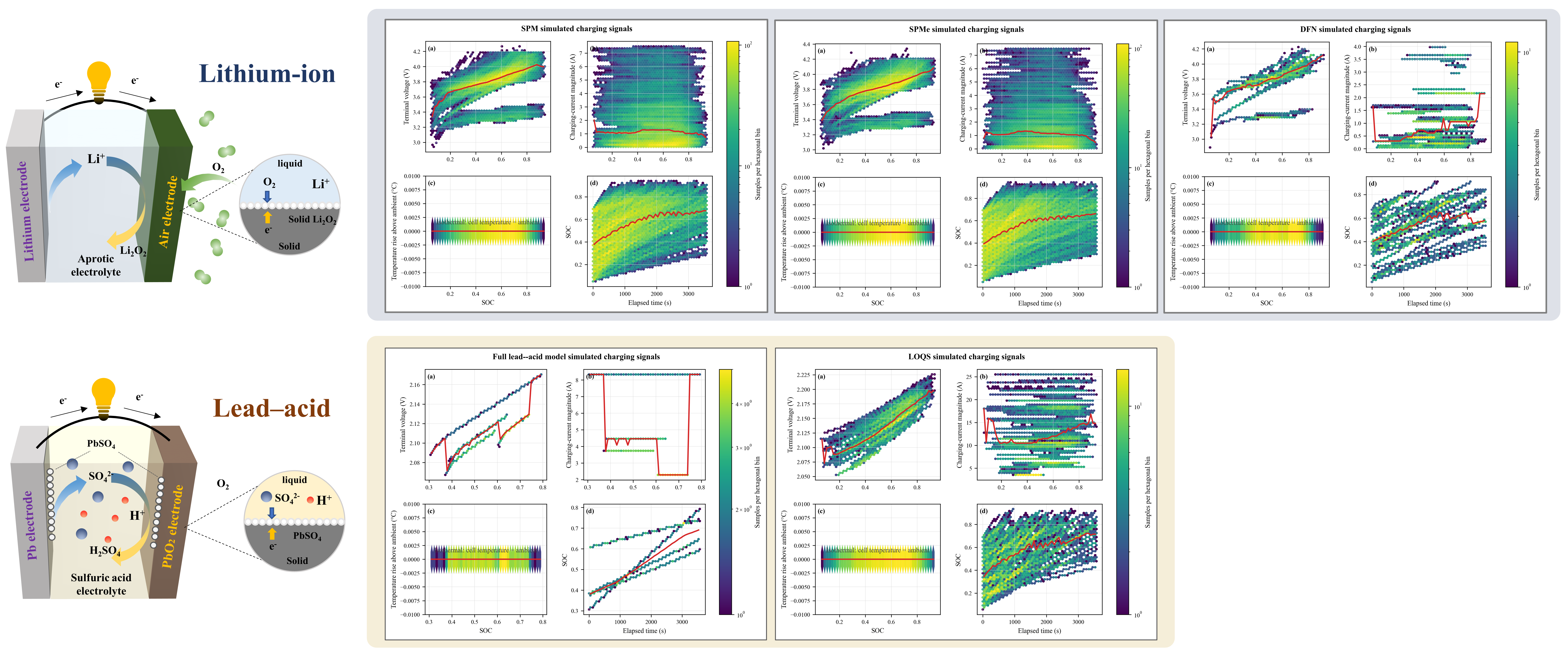}
    \caption{\textbf{Simulated charging-signal distributions for the five physics-based battery models.}                                                                                                                                                                                                  
    Results are presented for the SPM, SPMe, DFN, LOQS, and Full lead--acid                                                                                                                                                         
    models. For each model, the four panels show terminal voltage versus SOC,                                                                                                                                                       
    charging-current magnitude versus SOC, cell-temperature rise relative to                                                                                                                                                        
    ambient versus SOC, and SOC versus elapsed charging time, respectively.                                                                                                                                                         
    The hexagonal-bin color indicates the logarithmic sample density, whereas                                                                                                                                                       
    the red curve denotes the pooled binned median.} 
  \label{fig:simulation-signal-coverage}   
\end{figure*}

Fig.~\ref{fig:simulation-signal-coverage} visualizes the operating-state coverage of the trajectories generated by the five model families.                                                                                                                                                                
Each visualization jointly presents the terminal voltage, charging-current magnitude, cell-temperature rise, and SOC evolution.    
The terminal voltage generally increases with SOC for all five model families, while the absolute voltage range depends strongly on battery chemistry.                                                                                                                                                     
The lithium-ion models (SPM, SPMe, and DFN) primarily span approximately \(3.0\)--\(4.3\)~V, whereas the LOQS and Full lead--acid models operate near \(2.05\)--\(2.23\)~V. 
This representation shows both the physical response predicted by each model and the diversity of the observations subsequently converted into language-model training samples.                                                                                                                                                         
                                                                                                                                                      
\begin{table*}[t]
\centering
\caption{Physics-based battery models and parameterizations used to construct the simulation corpus. The equations summarize the principal physics retained by each model, while the numerical coefficients and constitutive relations are provided by the corresponding PyBaMM parameter sets.}
\label{tab:battery-models}

\footnotesize
\setlength{\tabcolsep}{4pt}
\renewcommand{\arraystretch}{1.12}

\setlength{\abovedisplayskip}{2pt}
\setlength{\belowdisplayskip}{2pt}
\setlength{\abovedisplayshortskip}{1pt}
\setlength{\belowdisplayshortskip}{1pt}

\begin{tabular}{
>{\raggedright\arraybackslash}p{0.070\textwidth}
>{\raggedright\arraybackslash}p{0.240\textwidth}
>{\raggedright\arraybackslash}p{0.635\textwidth}
}
\toprule
Model &
Chemistry \& parameters &
Resolution \& formulation \\
\midrule

SPM &
\makecell[l]{Lithium-ion;\\
10 parameter sets
\cite{ai2020electrochemical,chen2020development,ecker2015parameterization,
marquis2019asymptotic,mohtat2020differential,kim2011multi,
o2022lithium,o2022thermal,prada2013simplified,ramadass2004development}}
&
Single representative spherical particle is resolved in each electrode.
Solid-state diffusion and interfacial kinetics are retained, whereas
electrolyte spatial dynamics are neglected \cite{marquis2019asymptotic}.
The principal equations are
{\scriptsize
\[
\begin{aligned}
\frac{\partial c_{s,k}}{\partial t}
&=
\frac{1}{r^2}\frac{\partial}{\partial r}
\left(
D_{s,k}r^2\frac{\partial c_{s,k}}{\partial r}
\right), \quad
-D_{s,k}
\left.
\frac{\partial c_{s,k}}{\partial r}
\right|_{r=R_k}
&=
\frac{j_k}{F},\\
V
&=
U_p(c_{s,p}^{\mathrm{surf}})
-U_n(c_{s,n}^{\mathrm{surf}})
+\eta_p-\eta_n-I R_{\Omega}.
\end{aligned}
\]
}
\\

\addlinespace[2pt]

SPMe &
\makecell[l]{Lithium-ion;\\
10 parameter sets
\cite{ai2020electrochemical,chen2020development,ecker2015parameterization,
marquis2019asymptotic,mohtat2020differential,kim2011multi,
o2022lithium,o2022thermal,prada2013simplified,ramadass2004development}}
&
Extends the SPM by resolving through-cell electrolyte concentration and
potential dynamics, providing an intermediate-fidelity approximation
\cite{marquis2019asymptotic}. In addition to the SPM equations,
{\scriptsize
\[
\begin{aligned}
\varepsilon_e\frac{\partial c_e}{\partial t}
&=
-\frac{\partial N_e}{\partial x}
+\frac{1-t_+^0}{F}a_kj_k, \quad
N_e
&=
-D_e^{\mathrm{eff}}
\frac{\partial c_e}{\partial x}
+\frac{t_+^0}{F}i_e, \quad
\frac{\partial i_e}{\partial x}
&=
a_kj_k,
\end{aligned}
\]
}
with the corresponding electrolyte-polarization contribution included
in the terminal voltage.
\\

\addlinespace[2pt]

DFN &
\makecell[l]{Lithium-ion;\\
10 parameter sets
\cite{ai2020electrochemical,chen2020development,ecker2015parameterization,
marquis2019asymptotic,mohtat2020differential,kim2011multi,
o2022lithium,o2022thermal,prada2013simplified,ramadass2004development}}
&
Pseudo-two-dimensional porous-electrode model resolving both particle-scale
solid diffusion and electrode-scale electrolyte transport. The principal
electrochemical relations include
{\scriptsize
\[
\begin{aligned}
\frac{\partial}{\partial x}
\left(
\sigma_k^{\mathrm{eff}}
\frac{\partial\phi_{s,k}}{\partial x}
\right)
=
a_kj_k, \quad
\frac{\partial i_e}{\partial x}
&=
a_kj_k,\\
j_k
=
2j_{0,k}
\sinh\left(
\frac{F\eta_k}{2RT}
\right), \quad
\eta_k
&=
\phi_{s,k}-\phi_e-U_k,
\end{aligned}
\]
}
together with the solid-phase diffusion and electrolyte mass-balance
equations \cite{ha2024cobrapro}.
\\

\addlinespace[2pt]

LOQS &
\makecell[l]{Lead--acid;\\
Sulzer2019 parameter set \cite{sulzer2019faster}}
&
Leading-order quasi-static porous-electrode approximation for regimes
with relatively weak electrolyte concentration gradients. Starting from
{\scriptsize
\[
\begin{aligned}
\varepsilon\frac{\partial c_e}{\partial t}
&=
-\frac{\partial N_e}{\partial x}
+\frac{s_kj_k}{F}, \quad
\frac{\partial i_e}{\partial x}
&=
a_kj_k, \quad
i_s+i_e
&=
I,
\end{aligned}
\]
\[
\begin{aligned}
c_e(x,t)
&=
c_e^{(0)}(t)+O(\mathcal{C}_d), \quad
\phi_e(x,t)
&=
\phi_e^{(0)}(t)+O(\mathcal{C}_d),
\end{aligned}
\]
}
where $\mathcal{C}_d$ denotes the diffusional C-rate
\cite{sulzer2019faster}.
\\

\addlinespace[2pt]

Full &
\makecell[l]{Lead--acid;\\
Sulzer2019 parameter set \cite{sulzer2019faster}}
&
Spatially resolved porous-electrode model retaining electrolyte transport,
charge conservation, electrode kinetics, and porosity evolution:
{\scriptsize
\[
\begin{aligned}
\varepsilon\frac{\partial c_e}{\partial t}
&=
-\frac{\partial N_e}{\partial x}
+\frac{s_kj_k}{F},\quad
\frac{\partial i_e}{\partial x}
=
a_kj_k,\\
j_k
&=
2j_{0,k}
\sinh\left(
\frac{F\eta_k}{2RT}
\right),\quad
\frac{\partial\varepsilon}{\partial t}
=
-\beta_k j_k.
\end{aligned}
\]
}
This model provides the fully resolved lead--acid reference formulation
used in the corpus \cite{sulzer2019faster}.
\\

\bottomrule
\end{tabular}
\end{table*}

The resulting compatibility space contains 30 lithium-ion model--parameter-set combinations and two lead--acid combinations.
For each case, the simulator applies constant-current charging until the specified duration or the model-specific upper-voltage event is reached.  
The duration is bounded between 120 and 3600~s, and outputs are sampled every 30~s.  
Since the lead--acid interface does not accept the lithium-ion initial SOC argument, a lead--acid cell is first discharged from its initialized state at 0.5C to construct the requested starting condition and is then charged.

\subsection{Construction of Language-Model-Ready Simulation Corpora}
\label{subsec:llm-corpus-construction}

Each successful simulation produces an episode
\(\mathcal{E}_i=\{\mathbf{x}_{i,t}\}_{t=0}^{T_i}\).  At every sampled step,
the stored observation is
\begin{equation}
\begin{split}
\mathbf{x}_{i,t}=\{&m_i,t,z_t,z_{t+1},V_t,I_t,T_t,T_{\mathrm{amb},i},C_{\mathrm{rate},i},Q_i\},
\end{split}
\label{eq:stored-simulation-state}
\end{equation}
where \(m_i\) is the complete chemistry:model:parameter-set key, \(Q_i\) is nominal capacity.  
SOC is calculated by coulomb counting from the simulated current as Eq.\ref{eq:soc-coulomb-counting}.
\begin{equation}
 z_{t+1}=\operatorname{clip}_{[0,1]}\!\left[
 z_0-\frac{1}{Q_i}\int_0^{t+\Delta t}\frac{I(\tau)}{3600}\,\mathrm{d}\tau
 \right],
 \label{eq:soc-coulomb-counting}
\end{equation}

under PyBaMM's convention that discharge current is positive.  The integral
is evaluated by the trapezoidal rule.  Thus, although the underlying solvers
use richer internal electrochemical states, the language corpus exposes a
compact and model-agnostic interface consisting of directly actionable cell
features. 

The conversion from trajectories to text follows a causal next-state construction.  
For step \(t\), the input contains the current observation and at most the three preceding observations in Eq.\ref{eq:causal-corpus-pair}.
\begin{equation}
\begin{aligned}
 \mathbf{p}_{i,t}&=\operatorname{JSON}\!\left(
 \{\texttt{history}:\mathbf{x}_{i,t-3:t-1},
   \texttt{current}:\mathbf{x}_{i,t}\}\right),\\
 \mathbf{y}_{i,t}&=\operatorname{JSON}\!\left(
 \{\texttt{next\_soc}:z_{i,t+1}\}\right).
\end{aligned}
 \label{eq:causal-corpus-pair}
\end{equation}
Future samples are never included in the prompt.  
A representative serialized pair is just like profile below:
\begin{tcolorbox}[
    colback=white,
    colframe=black,
    boxrule=0.6pt,
    arc=0pt,
    outer arc=0pt,
    left=6pt,
    right=6pt,
    top=5pt,
    bottom=5pt
]
\scriptsize

Predict the immediate battery result from current causal state.

\medskip

\{
\textbf{``history'':}
\{
``step'':6,
``soc'':0.312,
``voltage\_v'':3.71,
``temperature\_c'':25.4
\},
\\[2pt]
\textbf{``current'':}
\{
``step'':7,
``soc'':0.319,
``voltage\_v'':3.73,
``temperature\_c'':25.5
\}
\}

\medskip

\{
\textbf{``next\_soc'':} 0.326
\}

\end{tcolorbox}

The compact JSON representation makes field boundaries explicit, is directly tokenizable, and can be parsed deterministically during validation.
The train, validation, and test partitions are formed by grouping on episode.
Consequently, all temporally adjacent samples from one simulated trajectory remain in the same partition, preventing future points from an episode from leaking into training through another split.  
This procedure converts heterogeneous electrochemical simulations into causal prompt--response pairs while retaining the physical provenance and episode boundaries required for reproducible language-model training.

    \section{Methodology}
\label{sec:methodology}

\begin{figure*}[htbp] 
  \centering
  \includegraphics[width=0.8\linewidth]{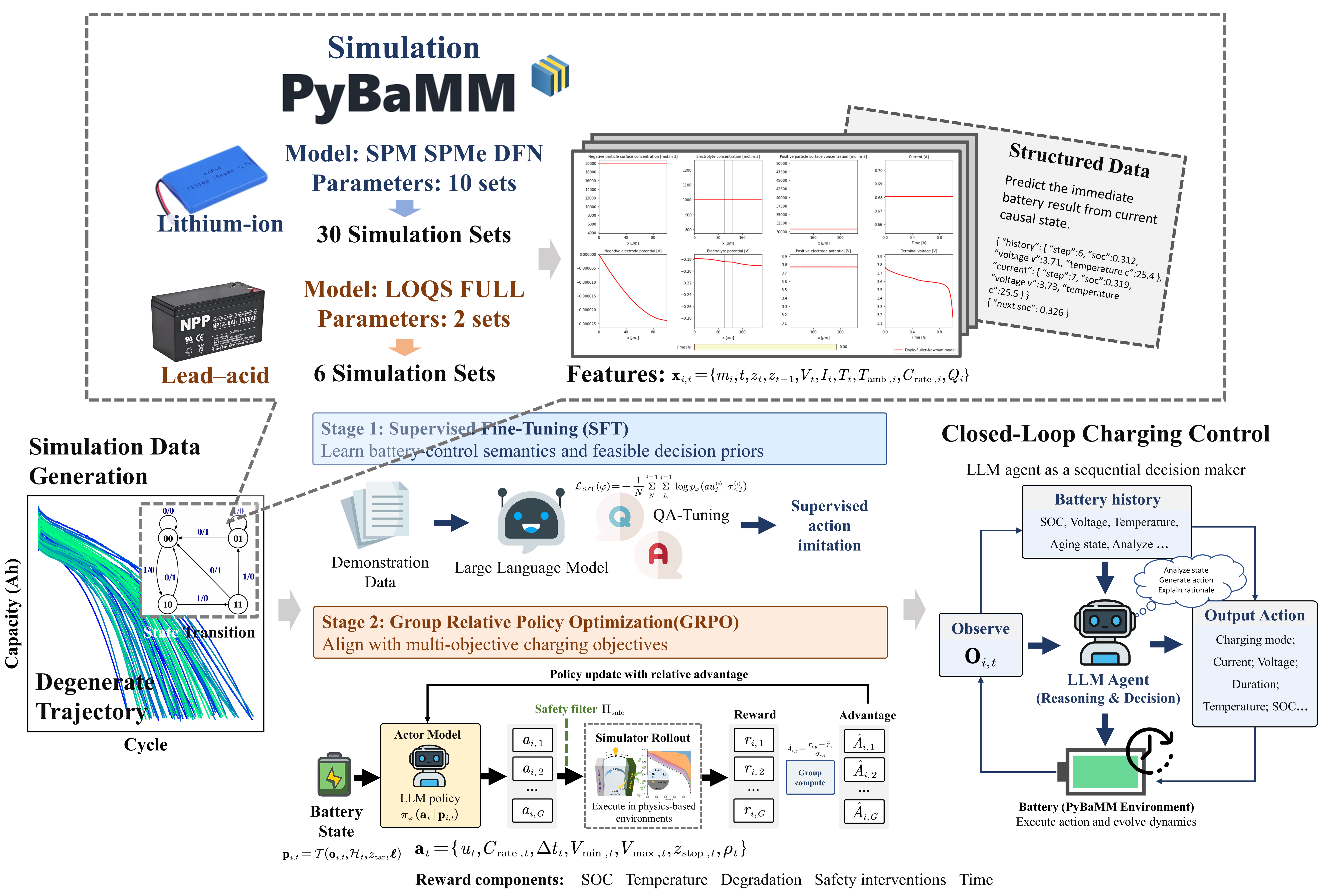}
\caption{Physics-grounded GRPO for adaptive battery fast charging. Starting from the SFT-initialized policy, the LLM agent samples multiple candidate actions for the same battery state. 
The actions are safety-filtered and evaluated through one-step PyBaMM rollouts, with rewards balancing SOC progress against temperature, degradation, safety intervention, and charging time. 
Group-relative advantages derived from these physics-based rewards are then used to update the policy toward safer and more effective charging decisions.}  
\label{fig:method}   
\end{figure*}

We formulate adaptive fast charging as a sequential decision-making problem in which a large language model (LLM) acts as the high-level policy of an intelligent battery-control agent in Fig\ref{fig:method} .
The proposed method follows a two-stage supervised fine-tuning and Group Relative Policy Optimization (SFT--GRPO) strategy.  
In the first stage, supervised fine-tuning teaches the LLM the structured battery-action protocol and initializes it with conservative expert charging behavior.  
In the second stage, GRPO improves the policy using one-step feedback from physics-based PyBaMM environments, thereby favoring actions that increase SOC rapidly while penalizing temperature, degradation, elapsed time, and safety-filter intervention.  
A deterministic safety layer remains outside the learned policy and constrains every action before it is applied to the battery model. 
This separation yields an agent that combines the contextual decision capability of an LLM with explicit
physical feedback and non-learned operational safeguards.

\subsection{State--Action Formulation}
\label{subsec:state-action-formulation}

Let \(\mathbf{o}_t\) denote the battery observation available at decision step \(t\).  
The implementation represents this observation as Eq.\ref{eq:agent-observation}.

\begin{equation}
 \mathbf{o}_{i,t}
 =\Pi_{\mathrm{causal}}(\mathbf{x}_{i,t})
 =\{t,z_t,V_t,I_t,T_t,T_{\mathrm{amb},i},Q_i\}.
 \label{eq:agent-observation}
\end{equation}
The future state \(z_{t+1}\) is excluded to prevent target leakage, while \(m_i\), \(s_i\), and \(q_i\) are provenance or quality-control metadata.
The scenario C-rate is not separately supplied because the instantaneous charging intensity is represented by \(I_t\) together with \(Q_i\).  
The software schema also carries fixed placeholders such as \texttt{soh=1} and zero-valued cumulative quantities when these fields are unavailable; 
they are not treated as informative features in the method.

The prompt combines \(\mathbf{o}_{i,t}\), up to three recent executed steps,
the target SOC \(z_{\mathrm{tar}}\), and the safety limits $\boldsymbol{\ell}$.  
The LLM policy is in Eq.\ref{eq:llm-policy}.
\begin{equation}
 \pi_{\varphi}(\mathbf{a}_t\mid\mathbf{p}_{i,t}),\qquad
 \mathbf{p}_{i,t}=\mathcal{T}\!\left(
 \mathbf{o}_{i,t},\mathcal{H}_t,z_{\mathrm{tar}},
 \boldsymbol{\ell}\right),
 \label{eq:llm-policy}
\end{equation}
where \(\mathcal{T}\) denotes textual serialization.  
Its output is a structured action in Eq.\ref{eq:structured-action}.
\begin{equation}
 \mathbf{a}_t=\{u_t,C_{\mathrm{rate},t},\Delta t_t,
 V_{\min,t},V_{\max,t},z_{\mathrm{stop},t},\rho_t\},
 \label{eq:structured-action}
\end{equation}
including the operating mode $u_t$, requested C-rate, duration, voltage and SOC conditions, and a short rationale $\rho_t$.  
The output JSON object is parsed into an executable battery action rather than used only as a textual recommendation.

\subsection{Stage I: Supervised Fine-Tuning}
\label{subsec:sft-stage}

\begin{table}[t]
\centering
\caption{Expert action initialization for SFT.}
\label{tab:sft-init}
\footnotesize
\setlength{\tabcolsep}{4pt}
\renewcommand{\arraystretch}{1.12}
\begin{tabular}{
>{\raggedright\arraybackslash}p{0.54\columnwidth}
>{\raggedright\arraybackslash}p{0.34\columnwidth}}
\toprule
State condition & Initialized action \\
\midrule
$z_t \ge z_{\mathrm{tar}}$
& \texttt{terminate} \\

$T_t \ge 55^{\circ}\mathrm{C}$ or $V_t \ge 4.17\,\mathrm{V}$
& \texttt{rest}(120 s) \\

$T_t \ge 42^{\circ}\mathrm{C}$ or $V_t \ge 4.05\,\mathrm{V}$
& \texttt{charge\_cc}(0.25C, 120 s) \\

$z_{\mathrm{tar}}-z_t \ge 0.35$ and $T_t \le 35^{\circ}\mathrm{C}$
& \texttt{charge\_cc}(0.8C, 120 s) \\

Otherwise
& \texttt{charge\_cc}(0.5C, 120 s) \\
\bottomrule
\end{tabular}
\end{table}

For every causal observation, a rule-based expert generates a safe target action in Table\ref{tab:sft-init} from the SOC deficit and the available voltage and temperature margins.  
This converts the simulated states into instruction--action pairs and teaches the LLM both the action protocol and a conservative initial charging policy.  
If
\(\boldsymbol{\tau}^{(i)}=(\tau_1^{(i)},\ldots,
\tau_{L_i}^{(i)})\) is the serialized prompt--action sequence, the LoRA-based
SFT stage minimizes the loss Eq.\ref{eq:sft-objective}.
\begin{equation}
 \mathcal{L}_{\mathrm{SFT}}(\varphi)
 =-\frac{1}{N}\sum_{i=1}^{N}\sum_{j=1}^{L_i}
 \log p_{\varphi}\!\left(	au_j^{(i)}\mid
 \tau_{<j}^{(i)}\right).
 \label{eq:sft-objective}
\end{equation}
Thus, SFT optimizes supervised action imitation rather than next-state
prediction: \(z_{t+1}\) remains a simulation outcome and is never exposed as
an input feature.

\subsection{Stage II: Physics-Grounded GRPO}
\label{subsec:grpo-stage}

Starting from the SFT policy, GRPO samples a group of candidate actions for the same battery state.  
Each valid candidate is safety-filtered and executed for one step in the corresponding simulation environment.  
For reward \(r_{i,g}\) of candidate \(g\) in group \(i\), the normalized relative advantage is Eq.\ref{eq:group-relative-advantage}
\begin{equation}
 \widehat{A}_{i,g}
 =\frac{r_{i,g}-\overline{r}_i}{\sigma_{r,i}},
 \qquad \sigma_{r,i}>0,
 \label{eq:group-relative-advantage}
\end{equation}
where \(\overline{r}_i\) and \(\sigma_{r,i}\) are the mean and standard deviation of the rewards for the same prompt.  
GRPO uses these within-prompt relative scores to reinforce completions that produce better charging outcomes than their peers, without requiring a separately trained value network.

The RL physical reward is shown in Eq.\ref{eq:grpo-reward}.
\begin{equation}
\begin{split}
 r_t={}&1+15(z_{t+1}-z_t)
 -0.2[T_{t+1}-25]_{+}
 -500[\Delta Q_{\mathrm{loss},t}]_{+}\\
 &-2N_{\mathrm{int},t}
 -0.01\frac{\Delta t_t}{60},
\end{split}
\label{eq:grpo-reward}
\end{equation}
where \([x]_{+}=\max(x,0)\) and \(N_{\mathrm{int},t}\) is the number of
safety interventions.  The reward balances SOC progress against absolute temperature, estimated capacity loss, dependence on safety correction, and elapsed charging time.  
Outputs that cannot be parsed, violate the action schema, or fail during physical rollout receive negative rewards and do not contribute a valid transition.  
Consequently, the two stages have distinct objectives: SFT learns protocol-compliant expert imitation, whereas GRPO optimizes physics-evaluated charging behavior.

\subsection{Safety-Constrained Deployment}
\label{subsec:safety-closed-loop}

Before execution, every LLM action is passed through deterministic rules that cap charge and discharge rates, restrict action duration and voltage bounds, and terminate control at the hard temperature limit.  
Malformed outputs are assigned categorical penalties before physical rollout, as summarized in Table~\ref{tab:invalid-penalties}.
Invalid JSON after two generation attempts is replaced by a conservative 0.2C, 60-s constant-current action.
\begin{table}[t]
\centering
\caption{Penalties for invalid agent outputs.}
\label{tab:invalid-penalties}
\footnotesize
\setlength{\tabcolsep}{5pt}
\renewcommand{\arraystretch}{1.10}

\begin{tabular}{
>{\raggedright\arraybackslash}p{0.45\columnwidth}
>{\centering\arraybackslash}p{0.15\columnwidth}}
\toprule
Invalid output & Penalty $r_t$ \\
\midrule
No JSON object found & $-12$ \\
Invalid JSON syntax & $-10$ \\
Invalid decoded action fields & $-8$ \\
Environment transition failure & $-6$ \\
\bottomrule
\end{tabular}
\end{table}

These penalties jointly enforce output syntax, action-schema validity, and physical executability.
Safety interventions are recorded and penalized by Eq.~\eqref{eq:grpo-reward}, encouraging the learned policy to satisfy the constraints before filtering.  
The closed-loop interaction is summarized by
\begin{equation}
 \mathbf{o}_{i,t}
 \xrightarrow{\pi_{\varphi}}\mathbf{a}_t
 \xrightarrow{\Pi_{\mathrm{safe}}}\widetilde{\mathbf{a}}_t
 \xrightarrow{\mathrm{PyBaMM}}\mathbf{o}_{i,t+1}.
 \label{eq:closed-loop-transition}
\end{equation}

    \section{Experiments}

\subsection{Experimental Setup}
\label{subsec:experimental-setup}

To evaluate the effectiveness and adaptability of the proposed LLM-based charging agent, experiments were conducted using four battery model--parameter-set combinations implemented in PyBaMM: DFN/Chen2020, SPM/Chen2020, SPMe/Chen2020, and SPM/Mohtat2020. 
Each configuration was evaluated under both isothermal and non-isothermal conditions, resulting in eight experimental scenarios. 
The initial cell and ambient temperatures were set to $25\,^{\circ}\mathrm{C}$. 
Under isothermal conditions, the temperature remained fixed, whereas non-isothermal simulations accounted for temperature evolution and propagated the resulting cell temperature between successive control actions.
All experiments adopted an initial SOC of 0.2, a charging target of 0.8, and a discharge target of 0.2. 
Cycling experiments were limited to 100 cycles, with a maximum of 32 control actions per charging or discharging phase. 
To ensure consistent operational constraints, all controllers shared a deterministic safety filter that limited charging and discharging rates to 1.5C and 2.0C, respectively, enforced a voltage range of 2.5--4.2\,V, and terminated operation when the cell temperature reached $60\,^{\circ}\mathrm{C}$.

The proposed agent was developed using Qwen3-4B-Instruct-2507 through a two-stage SFT--GRPO training procedure. 
The SFT stage employed 133,618 expert-labeled records generated from 32 battery model--parameter-set combinations to establish structured action generation and an initial charging policy. 
Subsequently, GRPO was initialized from the SFT adapter and optimized using up to 1,024 battery-state prompts, with candidate charging actions evaluated through single-step PyBaMM interactions.
The LLM operates within a hybrid discrete--continuous action space. 
The discrete action types include constant-current charging, constant-voltage charging, constant-current discharging, resting, and termination. 
Each action is further parameterized by numerical control variables, including the requested C-rate, action duration, lower and upper voltage limits, and stopping SOC. 
These decisions are generated as structured JSON outputs and translated into executable battery-control commands. 
Before execution, all actions are validated and processed by the shared deterministic safety filter to ensure compliance with the prescribed operational constraints.
The proposed agent was evaluated under three preference settings, $\beta\in\{0.01,0.10,0.40\}$, using a shared trained adapter. 
The preference parameter was specified through the inference prompt without additional fine-tuning, and model parameters remained fixed throughout evaluation.

Table~\ref{tab:control_protocols} summarizes the 16 controller configurations, comprising 13 baseline settings and three preference-conditioned LLM settings. 
The baselines cover conventional charging protocols, heuristic controllers, randomized control, and learning-based one-step optimization methods, providing comparisons across different control paradigms. 
All controllers were evaluated under identical battery configurations, thermal conditions, SOC targets, and safety constraints. 

\begin{table*}[!t]
\centering
\caption{Comparison of baseline charging strategies and the proposed LLM agent.}
\label{tab:control_protocols}

\scriptsize
\setlength{\tabcolsep}{6pt}
\renewcommand{\arraystretch}{1.08}

\begin{tabular}{@{}ll@{}}
\toprule
Controller & Charging strategy \\
\midrule

CCCV-0.5C / 1C / 2C*
& Constant-current charging followed by voltage-limited charging. \\

MSCC
& SOC-dependent charging at 1.2C, 0.8C, and 0.4C. \\

Pulse
& Three 1.2C charging steps followed by a rest period. \\

Thermal-rule
& Temperature-dependent C-rate: $\max\{0.2,1.2-0.035[T-25]_+\}$. \\

Fuzzy-like
& Heuristic C-rate adjustment based on SOC and temperature. \\

P-SOC
& Proportional SOC feedback (gain 3; 0.2--1.5C). \\

ECMS-like
& SOC-gap/rate and quadratic thermal-cost optimization over
  $\{0.3,0.6,0.9,1.2\}$C. \\

Random-safe
& Random C-rates (0.2--1.5C) and durations (60--180 s). \\

RF-LO / HGB-LO / MLP-LO
& Learned battery dynamics with one-step action optimization. \\

LLM-0.01 / 0.10 / 0.40
& SFT--GRPO policy with preference-conditioned charging decisions. \\

\bottomrule
\end{tabular}

\vspace{2pt}
\parbox{\textwidth}{
\scriptsize
\textit{Note:} CCCV-2C* is limited to 1.5C by the shared safety
filter. Fuzzy-like and ECMS-like denote heuristic implementations
rather than standard fuzzy-control or ECMS algorithms.
All controllers are subject to the same operational safety constraints.
}
\end{table*}

\subsection{Evaluation}

Performance is evaluated in terms of capacity retention, cycle life, and charging efficiency under isothermal and non-isothermal conditions. 
State of health (SOH) is the percentage of remaining capacity relative to its initial value, while $N_{80}$, $N_{50}$, and $N_{20}$ denote the first cycles at which SOH falls to or below the corresponding thresholds. 
Cumulative charging energy $E$ measures the energy processed during charging.
Table~\ref{tab:efficiency_beta_0_01} reports three additional metrics for non-isothermal charging at $\beta=0.01$. 
Charging-time reduction is defined as $100(1-\bar{t}_{\mathrm{LLM}}/\bar{t}_b)$, where $\bar{t}$ is the mean duration of successful charging stages. 
Matched-energy degradation reduction is $100(1-D_{\mathrm{LLM}}/D_b)$, where $D=\Delta Q/\Delta E$ represents capacity loss per unit charging energy (Ah/kWh), evaluated over the common interval $[E_{0.25},E_1]$ using interpolated capacities at its endpoints. 
Energy-rate gain is $100(P_{\mathrm{LLM}}/P_b-1)$, where $P=E_{\mathrm{charge}}/t_{\mathrm{charge}}$ represents charging-stage energy throughput per unit time over each controller's observed horizon. 

\begin{figure}[htbp] 
  \centering
  \includegraphics[width=\linewidth]{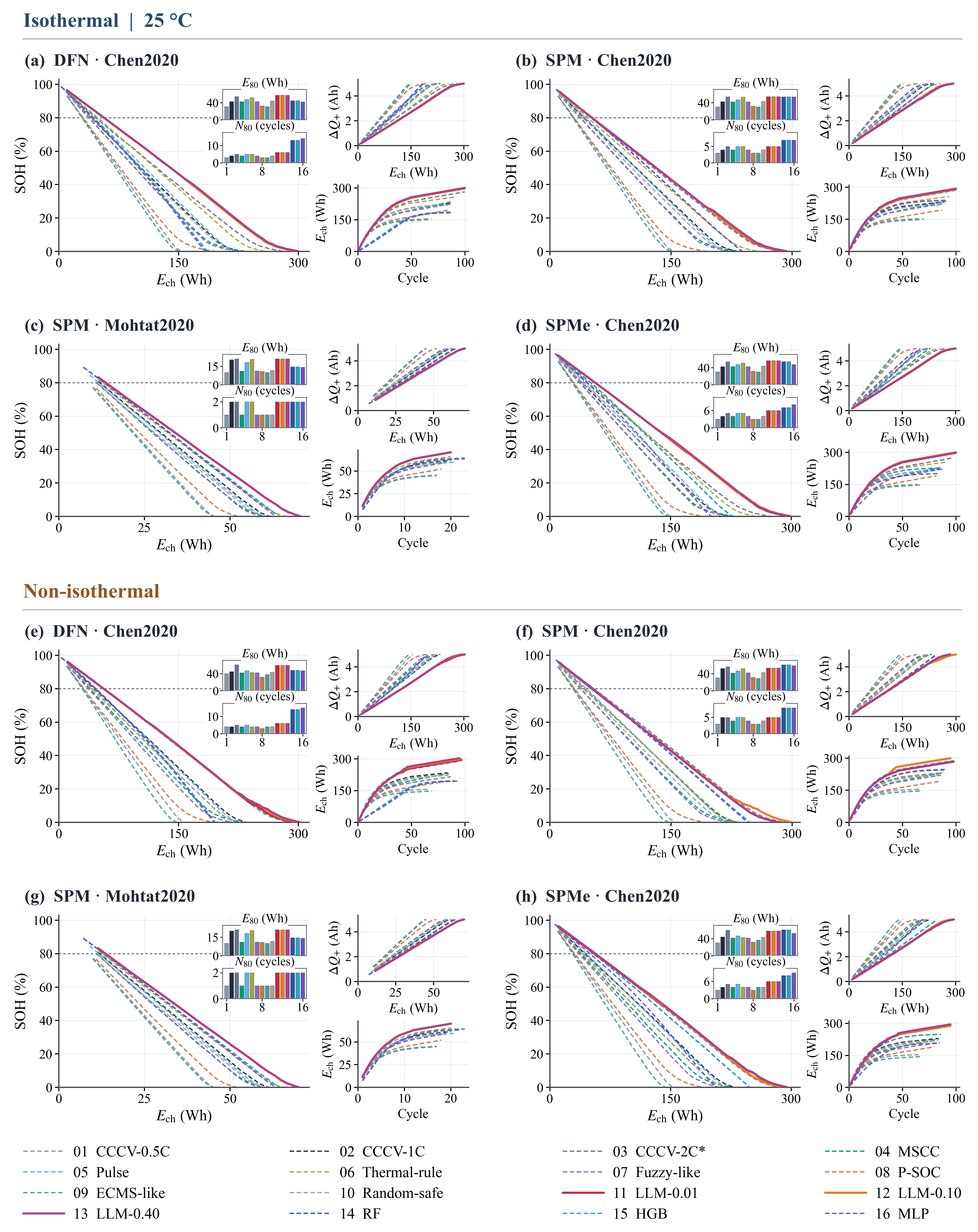}
    \caption{Lifetime and capacity-retention comparison across four battery model–parameter configurations under (a–d) isothermal (25 °C) and (e–h) non-isothermal conditions. 
    Each panel shows SOH versus cumulative charge-stage energy $E_{ch}$, with insets reporting E80 and N80 at the 80\% SOH threshold. 
    Right-hand plots show capacity loss versus energy and cumulative energy versus cycle. 
    Solid and dashed lines denote LLM strategies and baseline methods, respectively.
} 
  \label{result1}   
\end{figure}

\begin{table*}[!htbp]
\centering
\caption{Matched-energy capacity retention across thermal conditions.}
\label{tab:matched_energy_evidence}

\scriptsize
\setlength{\tabcolsep}{2pt}
\renewcommand{\arraystretch}{1.15}

\begin{tabular}{
>{\raggedright\arraybackslash}m{22mm}
>{\centering\arraybackslash}m{12mm}
>{\centering\arraybackslash}m{12mm}
>{\centering\arraybackslash}m{15mm}
>{\centering\arraybackslash}m{18mm}
>{\centering\arraybackslash}m{27mm}
>{\centering\arraybackslash}m{27mm}
>{\centering\arraybackslash}m{27mm}
}
\toprule

\shortstack[l]{Model\\Parameter set}
& Thermal
& $E^\ast$ (Wh)
& \shortstack{CCCV-1C\\SOH (\%)}
& \shortstack{Best\\non-LLM\\SOH (\%)}
& \shortstack{LLM ($\beta=.01$)\\SOH (\%)\\($\Delta$, pp)}
& \shortstack{LLM ($\beta=.10$)\\SOH (\%)\\($\Delta$, pp)}
& \shortstack{LLM ($\beta=.40$)\\SOH (\%)\\($\Delta$, pp)}
\\

\midrule

\multirow{2}{22mm}{\shortstack[l]{DFN\\Chen2020}}
& Iso. & 150.9 & 27.56 & 39.16
& \textbf{45.99 (+18.43)}
& \textbf{45.99 (+18.43)}
& \textbf{45.99 (+18.43)}
\\

& Noniso. & 147.7 & 33.66 & 46.01
& \textbf{46.73 (+13.08)}
& 46.55 (+12.89)
& \textbf{46.73 (+13.08)}
\\

\addlinespace[2pt]

\multirow{2}{22mm}{\shortstack[l]{SPM\\Chen2020}}
& Iso. & 149.9 & 29.97 & 41.78
& \textbf{43.08 (+13.11)}
& \textbf{43.08 (+13.11)}
& \textbf{43.08 (+13.11)}
\\

& Noniso. & 147.0 & 32.24 & \textbf{45.61}
& 43.87 (+11.63)
& 44.19 (+11.94)
& 43.87 (+11.63)
\\

\addlinespace[2pt]

\multirow{2}{22mm}{\shortstack[l]{SPMe\\Chen2020}}
& Iso. & 147.3 & 27.69 & 38.56
& 46.79 (+19.09)
& 47.08 (+19.39)
& \textbf{47.54 (+19.85)}
\\

& Noniso. & 143.7 & 34.37 & 47.08
& 47.90 (+13.53)
& 47.96 (+13.60)
& \textbf{48.49 (+14.13)}
\\

\addlinespace[2pt]

\multirow{2}{22mm}{\shortstack[l]{SPM\\Mohtat2020}}
& Iso. & 44.9 & 24.48 & \textbf{33.82}
& \textbf{33.82 (+9.34)}
& \textbf{33.82 (+9.34)}
& \textbf{33.82 (+9.34)}
\\

& Noniso. & 44.8 & 24.34 & \textbf{33.57}
& \textbf{33.57 (+9.22)}
& \textbf{33.57 (+9.22)}
& \textbf{33.57 (+9.22)}
\\

\bottomrule
\end{tabular}
\end{table*}

Under isothermal conditions, the capacity--energy trajectories in Fig.~\ref{result1} and the matched-energy results in Table~\ref{tab:matched_energy_evidence} consistently place the LLM among the strongest capacity-retention strategies.
For DFN/Chen2020, all three LLM preferences retain 45.99\% SOH at 150.9\,Wh, compared with 27.56\% for CCCV-1C and 39.16\% for the best non-LLM controller, CCCV-2C*.
The corresponding gains are 18.43 and 6.83\,pp, respectively.
For SPM/Chen2020, the LLM retains 43.08\% SOH, exceeding CCCV-1C by 13.11\,pp and the strongest baseline by 1.30\,pp.
The largest isothermal improvement occurs for SPMe/Chen2020: LLM SOH ranges from 46.79\% to 47.54\%, compared with 38.56\% for CCCV-2C*, with the $\beta=0.40$ setting achieving a reported 19.85\,pp gain over CCCV-1C.
For SPM/Mohtat2020, the LLM matches the strongest baseline at 33.82\% SOH while retaining a 9.34\,pp advantage over CCCV-1C.

\begin{figure*}[htbp] 
  \centering
  \includegraphics[width=0.9\linewidth]{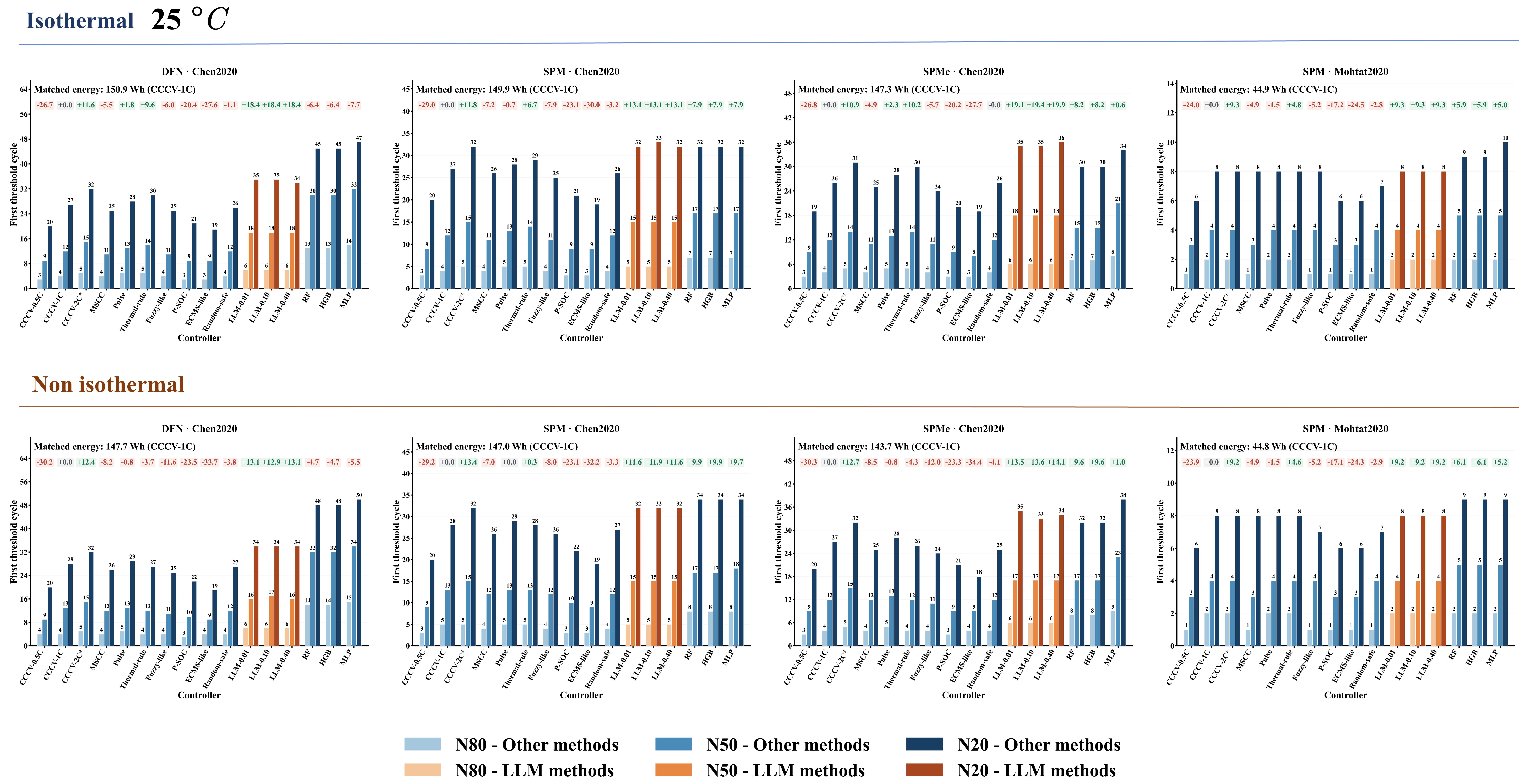}
    \caption{Cycle-based lifetime and matched-energy capacity retention. Bars show N80, N50, and N20; 
    annotations report SOH differences relative to CCCV-1C at equal cumulative charge-stage energy, in percentage points.} 
  \label{result2}   
\end{figure*}

The cycle-threshold results in Fig.~\ref{result2} complement these energy-based comparisons.
For isothermal SPMe/Chen2020, the LLM reaches $N_{80}=6$, $N_{50}=18$, and $N_{20}=35$--36 across the evaluated preferences, compared with 5, 14, and 31 for CCCV-2C*.
Its $N_{20}$ exceeds that of every non-LLM strategy in this configuration, indicating that the lifetime benefit persists into the deeper capacity-fade regime.
Together with the higher SOH at matched energy in Table~\ref{tab:matched_energy_evidence}, this result shows that the delayed threshold crossings are accompanied by improved capacity retention for a comparable amount of processed energy.

Under non-isothermal conditions, the LLM remains competitive in the joint comparison of capacity retention and energy throughput, although it does not maximize every individual metric.
For DFN/Chen2020, Fig.~\ref{result1} illustrates why recorded cycle count alone can be misleading: MLP reaches 80\% SOH at cycle 15, compared with cycle 6 for the LLM, but its cumulative charging energy at that threshold is only 46.80\,Wh, versus 59.39\,Wh for the LLM.
The latter is also close to the 60.73\,Wh achieved by CCCV-2C*.
Similarly, Fig.~\ref{result2} shows $N_{20}=48$--50 for RF, HGB, and MLP, exceeding the LLM's 34 cycles; nevertheless, their SOH at the matched cumulative energy is only 28.21\%--28.95\%, compared with 46.55\%--46.73\% for the LLM.
Table~\ref{tab:matched_energy_evidence} anchors this comparison at $E^\ast=147.7$\,Wh, where CCCV-1C and the strongest non-LLM baseline retain 33.66\% and 46.01\% SOH, respectively.
Thus, the additional cycles recorded by these learning-based baselines do not translate into superior capacity retention at equal energy throughput.
Under the same training objective, these complementary results support a better balance between capacity preservation and energy processing for the LLM, rather than an advantage confined to cycle count.

The remaining non-isothermal configurations clarify the extent of this benefit.
For SPMe/Chen2020, the LLM retains 47.90\%--48.49\% SOH at 143.7\,Wh, exceeding the strongest baseline's 47.08\% and achieving reported gains of 13.53--14.13\,pp over CCCV-1C.
For SPM/Mohtat2020, it matches the best baseline at 33.57\% SOH.
Non-isothermal SPM/Chen2020 is the exception: the LLM retains 43.87\%--44.19\% SOH, above CCCV-1C's 32.24\% but below CCCV-2C*'s 45.61\%.
Across the eight scenarios in Table~\ref{tab:matched_energy_evidence}, at least one LLM preference therefore exceeds the best non-LLM SOH in five scenarios and matches it in two.
Preference sensitivity is configuration-dependent: the three settings give identical reported SOH values in several scenarios, whereas $\beta=0.40$ provides the highest matched-energy SOH for SPMe/Chen2020 under both thermal conditions.

\begin{table}[!htbp]
\centering
\caption{Charging-time reduction, matched-energy degradation reduction,
and charge-stage energy-rate gain for LLM preference $\beta=0.01$.}
\label{tab:efficiency_beta_0_01}

\scriptsize
\setlength{\tabcolsep}{4pt}
\renewcommand{\arraystretch}{1.05}

\begin{tabular}{@{}lrrrrrr@{}}
\toprule
Comparator
& \multicolumn{3}{c}{DFN / Chen2020}
& \multicolumn{3}{c}{SPM / Chen2020} \\
\cmidrule(lr){2-4}
\cmidrule(l){5-7}

& \shortstack{Time\\red. (\%)}
& \shortstack{Degradation\\red. (\%)}
& \shortstack{Energy-rate\\gain (\%)}
& \shortstack{Time\\red. (\%)}
& \shortstack{Degradation\\red. (\%)}
& \shortstack{Energy-rate\\gain (\%)} \\
\midrule

CCCV-0.5C    & +65.8 & +44.3 & +207.0 & +54.1 & +42.2 & +203.7 \\
CCCV-1C      & +46.7 & +19.3 & +50.5  & +27.5 & +17.4 & +50.5  \\
CCCV-2C*     & +25.7 & +1.0  & +1.6   & +0.7  & -2.5  & -1.1   \\
MSCC         & +54.6 & +27.4 & +80.5  & +38.5 & +24.4 & +78.7  \\
Pulse        & +29.5 & +21.7 & +41.3  & +6.3  & +19.0 & +40.7  \\
Thermal-rule & +47.1 & +23.1 & +60.6  & +25.8 & +16.7 & +47.8  \\
Fuzzy-like   & +48.6 & +30.8 & +75.8  & +30.8 & +25.7 & +63.3  \\
P-SOC        & +56.1 & +39.2 & +115.8 & +40.0 & +37.0 & +112.4 \\
ECMS-like    & +60.5 & +45.6 & +232.6 & +50.7 & +43.0 & +221.6 \\
Random-safe  & +53.7 & +24.1 & +75.9  & +43.8 & +21.5 & +74.7  \\
RF           & -2.1  & +28.1 & +123.8 & -46.4 & +6.0  & -0.8   \\
HGB          & -2.1  & +28.1 & +123.8 & -46.4 & +6.0  & -0.8   \\
MLP          & -1.7  & +29.0 & +133.5 & -44.2 & +6.4  & +2.8   \\

\midrule

Comparator
& \multicolumn{3}{c}{SPMe / Chen2020}
& \multicolumn{3}{c}{SPM / Mohtat2020} \\
\cmidrule(lr){2-4}
\cmidrule(l){5-7}

& \shortstack{Time\\red. (\%)}
& \shortstack{Degradation\\red. (\%)}
& \shortstack{Energy-rate\\gain (\%)}
& \shortstack{Time\\red. (\%)}
& \shortstack{Degradation\\red. (\%)}
& \shortstack{Energy-rate\\gain (\%)} \\
\midrule

CCCV-0.5C    & +68.6 & +45.4 & +208.1 & +47.6 & +31.9 & +204.4 \\
CCCV-1C      & +50.6 & +20.2 & +51.6  & +27.5 & +11.9 & +50.9  \\
CCCV-2C*     & +31.4 & +1.3  & +2.0   & +0.0  & +0.0  & +0.0   \\
MSCC         & +58.3 & +28.9 & +82.1  & +40.8 & +16.2 & +79.5  \\
Pulse        & +36.0 & +22.7 & +41.8  & -9.6  & +13.9 & +42.0  \\
Thermal-rule & +52.0 & +24.6 & +62.3  & +14.9 & +6.1  & +25.8  \\
Fuzzy-like   & +53.5 & +32.1 & +78.0  & +34.7 & +17.1 & +65.3  \\
P-SOC        & +60.0 & +40.0 & +115.5 & +45.0 & +26.5 & +120.8 \\
ECMS-like    & +64.2 & +46.7 & +236.4 & +46.7 & +31.2 & +202.6 \\
Random-safe  & +58.9 & +25.6 & +74.7  & +38.0 & +12.4 & +70.9  \\
RF           & +9.3  & +9.0  & +3.6   & -82.6 & +5.3  & +3.9   \\
HGB          & +9.3  & +9.0  & +3.6   & -82.6 & +5.3  & +3.9   \\
MLP          & -2.0  & +23.0 & +65.9  & -44.0 & +7.7  & +17.3  \\

\bottomrule
\end{tabular}
\end{table}

The efficiency results in Table~\ref{tab:efficiency_beta_0_01} demonstrate that the LLM achieves improved capacity retention while maintaining competitive charging efficiency. 
Compared with CCCV-1C, the LLM ($\beta=0.01$) reduces charging time by 27.5\%--50.6\% and matched-energy degradation by 11.9\%--20.2\%, while increasing the charge-stage energy rate by 50.5\%--51.6\% across four non-isothermal configurations. 
Similar improvements are observed against most conventional and heuristic baselines. 
Compared with RF, HGB, and MLP, the LLM consistently reduces matched-energy degradation across all configurations, with reductions of 28.1\%--29.0\% for DFN/Chen2020, despite configuration-dependent trade-offs in charging time and energy rate. 
Overall, these results indicate that the proposed agent provides a favorable balance between degradation mitigation and charging efficiency, rather than uniformly outperforming all baselines on every metric.

Overall, the proposed LLM policy combines strong matched-energy capacity retention with competitive energy throughput and, in selected configurations, delayed capacity-fade thresholds.
It improves matched-energy SOH over CCCV-1C in all eight scenarios and reaches or exceeds the strongest non-LLM result in seven, while simultaneously improving all three reported non-isothermal efficiency metrics relative to CCCV-1C.
Its principal advantage is the balance achieved across degradation, energy processing, and charging performance under shared operational constraints, rather than the largest recorded cycle count in every case.
These conclusions describe the evaluated simulation trajectories; each controller--configuration pair is represented by one trajectory, so the reported differences are descriptive rather than claims of statistical significance.

    \section{Discussion}

\subsection{Action analysis}

\begin{figure*}[htbp] 
  \centering
  \includegraphics[width=0.8\linewidth]{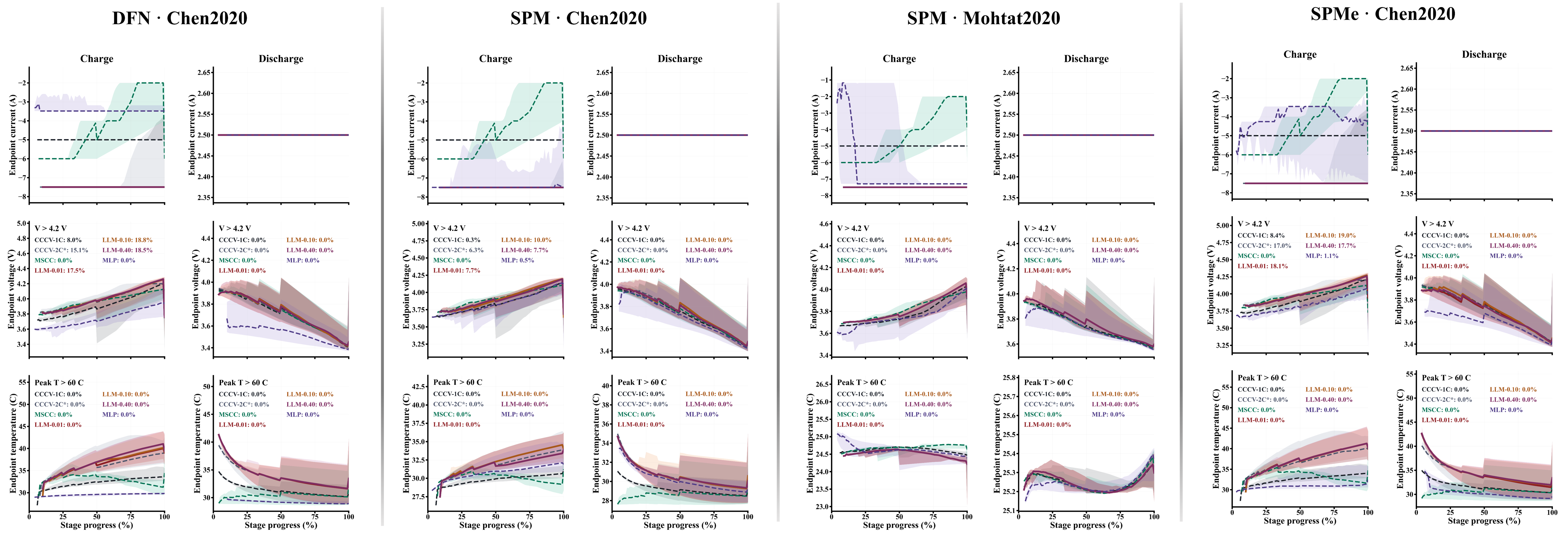}
    \caption{Charge and discharge endpoint current, voltage, and temperature versus normalized stage progress. 
    Lines and shading indicate cycle medians and 10th–90th percentiles. 
    Annotations report action-level voltage and peak-temperature exceed rates.
} 
  \label{result3}   
\end{figure*}

Figure~\ref{result3} examines the control behavior underlying the capacity-retention and lifetime results in Figs.~\ref{result1} and~\ref{result2}.
The curves summarize charge and discharge endpoint currents, voltages, and temperatures over normalized stage progress, with cycle medians and 10th--90th percentile bands indicating typical behavior and its variability.
For DFN/Chen2020, the LLM maintains a charging-current magnitude of approximately 7.5\,A over most of the charging stage, approaching the aggressive charging behavior of the high-rate baseline.
This current is higher than those of CCCV-1C and MLP and differs from the progressively decreasing current adopted by MSCC.
The resulting policy therefore makes substantial use of the permitted charging intensity.

The action-level exceedance rates in Fig.~\ref{result3} also clarify the role of the deterministic safety layer.
Across the four lithium-ion configurations, no recorded LLM action has a sampled peak temperature above $60\,^{\circ}\mathrm{C}$.
However, its voltage-exceedance rate is higher than those of MSCC, MLP, and the CCCV baselines in some configurations.
These observations support the performance competitiveness of the learned policy while underscoring the necessity of an explicit, controllable safety layer around its decisions.
Constraining a requested action does not by itself establish that every point on the resulting physical trajectory remains within the prescribed limits, and sampled temperature results should be interpreted at the resolution of the recorded observations.
The safety layer should therefore remain independent of the policy and retain authority to limit or reject actions and terminate operation.
Further refinement should focus on voltage-aware action screening and constraint monitoring during execution, so that the capacity-retention and throughput benefits are accompanied by more consistent voltage compliance.

\subsection{Inference Efficiency and Industrial Deployment}

Computational overhead is another consideration when transferring the proposed framework to industrial energy-storage systems.
In our implementation, Qwen3-4B-Instruct-2507 running on an NVIDIA RTX 4090D requires approximately 4.12\,s to infer one control action.
This latency was acceptable for the present experiments, in which the policy achieved competitive charging and capacity-retention performance.
At larger deployment scales, however, repeated inference across many independently controlled batteries would impose additional computing requirements and deployment costs, while inference latency would constrain the achievable decision frequency.
Reducing the computational burden of each decision is therefore an important step toward practical deployment.

Battery charging and discharging control offers an opportunity to specialize the inference interface.
The controller consumes structured battery-state information and must return an executable decision, including an operating mode and numerical control parameters; generating an extended natural-language response is not essential to this control loop.
Although the task also involves continuous parameter selection and sequential consequences, parts of it can be formulated as state assessment, candidate-action selection, and scoring.
This motivates investigating Jev, TypeSafe's decision-oriented System One model, which returns predefined typed decisions and associated probabilities.
TypeSafe reports approximately 193.6-fold faster execution and 444.6-fold lower cost in its selected System One workflow evaluations\cite{duan2026general}.
Its published input price is US\$0.042 per million tokens, with no output-token charge\cite{jiang2026jev}.
These vendor-reported workflow comparisons indicate the potential efficiency of specialized decision models.

Future work will investigate lighter and more efficient controllers for specialized industrial tasks based on Jev's decision-oriented approach.
One direction is to express battery-state assessment and the selection of feasible charging actions as typed decision queries, while assigning continuous control parameters through bounded optimization or a compact numerical policy.
This separation would allow a specialized decision model to handle state-dependent choices while preserving the explicit safety layer and the multi-objective balance established in the present framework.
The aim is to retain the benefits of adaptive charging with a smaller inference burden, making such control more practical for widely distributed energy-storage applications.

    \section{Conclusion}
\label{sec:conclusion}

Conventional battery charging strategies often rely on fixed protocols or condition-specific optimization, limiting their adaptability to heterogeneous battery characteristics, varying operating conditions, and evolving aging states. 
Balancing charging efficiency, energy throughput, and long-term degradation therefore remains a significant challenge in practical battery management. 
To address these limitations, this study developed an LLM-empowered controller for adaptive fast-charging control. A two-stage SFT--GRPO framework was proposed, where SFT establishes feasible charging behaviors from large-scale simulation data, while GRPO further improves control decisions through physics-based feedback. 
By integrating the LLM policy with the PyBaMM environment and a deterministic safety filter, the proposed framework enables structured and state-dependent closed-loop charging control.
Extensive simulations across different battery models, parameter configurations, and thermal conditions demonstrate the effectiveness of the proposed approach. 
Compared with conventional charging protocols and learning-based optimization methods, the proposed method generally achieves better capacity retention under matched energy-throughput conditions while maintaining competitive charging efficiency. 
The results further highlight its potential to adapt charging decisions to heterogeneous battery dynamics and balance short-term charging demands against long-term degradation, although the relative advantages vary across configurations.

Overall, this study demonstrates the potential of LLM-based control approaches to extend intelligent decision making from digital environments to the control of physical energy systems. 
Future research will focus on improving inference efficiency and safety-constraint enforcement, together with hardware-in-the-loop and real-cell validation, to facilitate the practical deployment of LLM-driven battery management.

        \fontsize{8}{9}\selectfont{
    \scriptsize{\bibliographystyle{ieeetr}
    \bibliography{main.bib}}}

\begin{thebibliography}{10}

\bibitem{chan2026universal}
J.~Chan, H.~Wang, Z.~Chen, H.~Pan, W.~Wu, Z.~Wang, E.~Pan, M.~Xie, and L.~Xi, ``Universal battery capacity degradation forecasting driven by foundation models across diverse chemistries and conditions,'' {\em Advanced Science}, p.~e77261, 2026.

\bibitem{tomaszewska2019lithium}
A.~Tomaszewska, Z.~Chu, X.~Feng, S.~O'kane, X.~Liu, J.~Chen, C.~Ji, E.~Endler, R.~Li, L.~Liu, {\em et~al.}, ``Lithium-ion battery fast charging: A review,'' {\em ETransportation}, vol.~1, p.~100011, 2019.

\bibitem{park2020reinforcement}
S.~Park, A.~Pozzi, M.~Whitmeyer, H.~Perez, W.~T. Joe, D.~M. Raimondo, and S.~Moura, ``Reinforcement learning-based fast charging control strategy for li-ion batteries,'' in {\em 2020 IEEE Conference on Control Technology and Applications (CCTA)}, pp.~100--107, IEEE, 2020.

\bibitem{ji2025design}
G.~Ji, L.~He, T.~Wu, and G.~Cui, ``The design of fast charging strategy for lithium-ion batteries and intelligent application: A comprehensive review,'' {\em Applied Energy}, vol.~377, p.~124538, 2025.

\bibitem{jiang2025fast}
B.~Jiang, Y.~Wang, Z.~Ma, and Q.~Lu, ``Fast charging of lithium-ion batteries using deep bayesian optimization with recurrent neural network,'' {\em IEEE Transactions on Automation Science and Engineering}, vol.~22, pp.~13150--13160, 2025.

\bibitem{trimboli2015charging}
M.~S. Trimboli and M.~A. Xavier, ``Charging strategies for lithium-ion batteries using electrochemical-based reduced-order models and predictive control,'' in {\em Proceedings of the Advanced Automotive Battery Conference (AABC), Detroit, MI, USA}, pp.~15--19, 2015.

\bibitem{gao2019health}
Y.~Gao, X.~Zhang, B.~Guo, C.~Zhu, J.~Wiedemann, L.~Wang, and J.~Cao, ``Health-aware multiobjective optimal charging strategy with coupled electrochemical-thermal-aging model for lithium-ion battery,'' {\em IEEE Transactions on industrial Informatics}, vol.~16, no.~5, pp.~3417--3429, 2019.

\bibitem{liu2018multi}
K.~Liu, K.~Li, H.~Ma, J.~Zhang, and Q.~Peng, ``Multi-objective optimization of charging patterns for lithium-ion battery management,'' {\em Energy Conversion and Management}, vol.~159, pp.~151--162, 2018.

\bibitem{liu2017constrained}
K.~Liu, K.~Li, and C.~Zhang, ``Constrained generalized predictive control of battery charging process based on a coupled thermoelectric model,'' {\em Journal of Power Sources}, vol.~347, pp.~145--158, 2017.

\bibitem{guo2011single}
M.~Guo, G.~Sikha, and R.~E. White, ``Single-particle model for a lithium-ion cell: Thermal behavior,'' {\em Journal of The Electrochemical Society}, vol.~158, no.~2, pp.~A122--A132, 2011.

\bibitem{kong2020pseudo}
X.~Kong, G.~L. Plett, M.~S. Trimboli, Z.~Zhang, D.~Qiao, T.~Zhao, and Y.~Zheng, ``Pseudo-two-dimensional model and impedance diagnosis of micro internal short circuit in lithium-ion cells,'' {\em Journal of Energy Storage}, vol.~27, p.~101085, 2020.

\bibitem{zhang2017charging}
C.~Zhang, J.~Jiang, Y.~Gao, W.~Zhang, Q.~Liu, and X.~Hu, ``Charging optimization in lithium-ion batteries based on temperature rise and charge time,'' {\em Applied energy}, vol.~194, pp.~569--577, 2017.

\bibitem{attia2020closed}
P.~M. Attia, A.~Grover, N.~Jin, K.~A. Severson, T.~M. Markov, Y.-H. Liao, M.~H. Chen, B.~Cheong, N.~Perkins, Z.~Yang, {\em et~al.}, ``Closed-loop optimization of fast-charging protocols for batteries with machine learning,'' {\em Nature}, vol.~578, no.~7795, pp.~397--402, 2020.

\bibitem{jiang2022fast}
B.~Jiang, M.~D. Berliner, K.~Lai, P.~A. Asinger, H.~Zhao, P.~K. Herring, M.~Z. Bazant, and R.~D. Braatz, ``Fast charging design for lithium-ion batteries via bayesian optimization,'' {\em Applied Energy}, vol.~307, p.~118244, 2022.

\bibitem{zhao2024reinforcement}
Z.~Zhao, C.~K. Lee, X.~Yan, and H.~Wang, ``Reinforcement learning for electric vehicle charging scheduling: A systematic review,'' {\em Transportation Research Part E: Logistics and Transportation Review}, vol.~190, p.~103698, 2024.

\bibitem{zhang2025fast}
Z.~Zhang, T.~Guo, Y.~Liu, X.~Pang, and Z.~Zheng, ``Fast-charging optimization method for lithium-ion battery packs based on deep deterministic policy gradient algorithm,'' {\em Batteries}, vol.~11, no.~5, p.~199, 2025.

\bibitem{hao2023adaptive}
Y.~Hao, Q.~Lu, X.~Wang, and B.~Jiang, ``Adaptive model-based reinforcement learning for fast-charging optimization of lithium-ion batteries,'' {\em IEEE Transactions on Industrial Informatics}, vol.~20, no.~1, pp.~127--137, 2023.

\bibitem{wu2023bloomberggpt}
S.~Wu, O.~Irsoy, S.~Lu, V.~Dabravolski, M.~Dredze, S.~Gehrmann, P.~Kambadur, D.~Rosenberg, and G.~Mann, ``Bloomberggpt: A large language model for finance,'' {\em arXiv preprint arXiv:2303.17564}, 2023.

\bibitem{lewkowycz2022solving}
A.~Lewkowycz, A.~Andreassen, D.~Dohan, E.~Dyer, H.~Michalewski, V.~Ramasesh, A.~Slone, C.~Anil, I.~Schlag, T.~Gutman-Solo, {\em et~al.}, ``Solving quantitative reasoning problems with language models,'' {\em Advances in neural information processing systems}, vol.~35, pp.~3843--3857, 2022.

\bibitem{azerbayev2024llemma}
Z.~Azerbayev, H.~Schoelkopf, K.~Paster, M.~Dos~Santos, S.~McAleer, Q.~Jiang, J.~Deng, S.~R. Biderman, and S.~Welleck, ``Llemma: An open language model for mathematics,'' in {\em International Conference on Learning Representations}, vol.~2024, pp.~40622--40649, 2024.

\bibitem{yao2022react}
S.~Yao, J.~Zhao, D.~Yu, N.~Du, I.~Shafran, K.~Narasimhan, and Y.~Cao, ``React: Synergizing reasoning and acting in language models,'' {\em arXiv preprint arXiv:2210.03629}, 2022.

\bibitem{wang2023voyager}
G.~Wang, Y.~Xie, Y.~Jiang, A.~Mandlekar, C.~Xiao, Y.~Zhu, L.~Fan, and A.~Anandkumar, ``Voyager: An open-ended embodied agent with large language models,'' {\em arXiv preprint arXiv:2305.16291}, 2023.

\bibitem{long2024llms}
L.~Long, R.~Wang, R.~Xiao, J.~Zhao, X.~Ding, G.~Chen, and H.~Wang, ``On llms-driven synthetic data generation, curation, and evaluation: A survey,'' in {\em Findings of the Association for Computational Linguistics: ACL 2024}, pp.~11065--11082, 2024.

\bibitem{sulzer2021python}
V.~Sulzer, S.~G. Marquis, R.~Timms, M.~Robinson, and S.~J. Chapman, ``Python battery mathematical modelling (pybamm),'' {\em Journal of Open Research Software}, vol.~9, no.~1, pp.~14--14, 2021.

\bibitem{shao2024deepseekmath}
Z.~Shao, P.~Wang, Q.~Zhu, R.~Xu, J.~Song, X.~Bi, H.~Zhang, M.~Zhang, Y.~Li, Y.~Wu, {\em et~al.}, ``Deepseekmath: Pushing the limits of mathematical reasoning in open language models,'' {\em arXiv preprint arXiv:2402.03300}, 2024.

\bibitem{ai2020electrochemical}
W.~Ai, L.~Kraft, J.~Sturm, A.~Jossen, and B.~Wu, ``Electrochemical thermal-mechanical modelling of stress inhomogeneity in lithium-ion pouch cells,'' {\em Journal of The Electrochemical Society}, vol.~167, no.~1, p.~013512, 2020.

\bibitem{chen2020development}
C.-H. Chen, F.~Brosa~Planella, K.~O’regan, D.~Gastol, W.~D. Widanage, and E.~Kendrick, ``Development of experimental techniques for parameterization of multi-scale lithium-ion battery models,'' {\em Journal of The Electrochemical Society}, vol.~167, no.~8, p.~080534, 2020.

\bibitem{ecker2015parameterization}
M.~Ecker, T.~K.~D. Tran, P.~Dechent, S.~K{\"a}bitz, A.~Warnecke, and D.~U. Sauer, ``Parameterization of a physico-chemical model of a lithium-ion battery: I. determination of parameters,'' {\em Journal of The Electrochemical Society}, vol.~162, no.~9, pp.~A1836--A1848, 2015.

\bibitem{marquis2019asymptotic}
S.~G. Marquis, V.~Sulzer, R.~Timms, C.~P. Please, and S.~J. Chapman, ``An asymptotic derivation of a single particle model with electrolyte,'' {\em Journal of The Electrochemical Society}, vol.~166, no.~15, pp.~A3693--A3706, 2019.

\bibitem{mohtat2020differential}
P.~Mohtat, S.~Lee, V.~Sulzer, J.~B. Siegel, and A.~G. Stefanopoulou, ``Differential expansion and voltage model for li-ion batteries at practical charging rates,'' {\em Journal of The Electrochemical Society}, vol.~167, no.~11, p.~110561, 2020.

\bibitem{kim2011multi}
G.-H. Kim, K.~Smith, K.-J. Lee, S.~Santhanagopalan, and A.~Pesaran, ``Multi-domain modeling of lithium-ion batteries encompassing multi-physics in varied length scales,'' {\em Journal of the electrochemical society}, vol.~158, no.~8, pp.~A955--A969, 2011.

\bibitem{o2022lithium}
S.~E. O'Kane, W.~Ai, G.~Madabattula, D.~Alonso-Alvarez, R.~Timms, V.~Sulzer, J.~S. Edge, B.~Wu, G.~J. Offer, and M.~Marinescu, ``Lithium-ion battery degradation: how to model it,'' {\em Physical Chemistry Chemical Physics}, vol.~24, no.~13, pp.~7909--7922, 2022.

\bibitem{o2022thermal}
K.~O'Regan, F.~B. Planella, W.~D. Widanage, and E.~Kendrick, ``Thermal-electrochemical parameters of a high energy lithium-ion cylindrical battery,'' {\em Electrochimica Acta}, vol.~425, p.~140700, 2022.

\bibitem{prada2013simplified}
E.~Prada, D.~Di~Domenico, Y.~Creff, J.~Bernard, V.~Sauvant-Moynot, and F.~Huet, ``A simplified electrochemical and thermal aging model of lifepo4-graphite li-ion batteries: power and capacity fade simulations,'' {\em Journal of The Electrochemical Society}, vol.~160, no.~4, pp.~A616--A628, 2013.

\bibitem{ramadass2004development}
P.~Ramadass, B.~Haran, P.~M. Gomadam, R.~White, and B.~N. Popov, ``Development of first principles capacity fade model for li-ion cells,'' {\em Journal of the Electrochemical Society}, vol.~151, no.~2, pp.~A196--A203, 2004.

\bibitem{ha2024cobrapro}
S.~Ha and S.~Onori, ``Cobrapro: An open-source software for the doyle-fuller-newman model with co-simulation parameter optimization framework,'' {\em Journal of The Electrochemical Society}, vol.~171, no.~9, p.~090522, 2024.

\bibitem{sulzer2019faster}
V.~Sulzer, S.~J. Chapman, C.~P. Please, D.~A. Howey, and C.~W. Monroe, ``Faster lead-acid battery simulations from porous-electrode theory: Part ii. asymptotic analysis,'' {\em Journal of The Electrochemical Society}, vol.~166, no.~12, pp.~A2372--A2382, 2019.

\bibitem{duan2026general}
F.~Duan, J.~Lin, J.~Wang, J.~Xiang, J.~Wu, X.~Zhang, H.~Yan, S.~Wang, and Z.~Wei, ``General decision models: Benchmarking and insights beyond jev,'' {\em arXiv preprint arXiv:2610.03935}, 2026.

\bibitem{jiang2026jev}
D.~Jiang, Y.~Li, and B.~Li, ``Jev-mem: System-one-controlled agentic memory for efficient ai agents,'' {\em arXiv preprint arXiv:2609.23986}, 2026.

\end{thebibliography}

\end{document}